\documentclass[conference]{IEEEtran}
\usepackage{cite}
\usepackage[colorlinks=true,linkcolor=blue,citecolor=blue,urlcolor=blue]{hyperref}
\makeatletter
\AtBeginDocument{%
    \catcode`_=12
    \begingroup\lccode`~=`_
    \lowercase{\endgroup\let~}\sb
    \mathcode`_="8000
    \immediate\write\@auxout{\catcode`_=12 }%
    \immediate\write\@auxout{\catcode`^=12 }%
}
\makeatother
\usepackage{caption} 
\usepackage{graphicx}
\usepackage{algorithm}
\usepackage{algpseudocode}
\usepackage{subfigure} 
\usepackage{amsmath,amssymb,amsfonts}
\usepackage{textcomp}
\usepackage{booktabs}  
\usepackage{multirow}
\usepackage{xcolor}
\usepackage{makecell}  
\usepackage{bm}
\usepackage{bookmark}
\usepackage{dblfloatfix}
\usepackage{capt-of}
\usepackage[top=0.75in, bottom=0.75in, left=0.75in, right=0.75in]{geometry}
\newcommand{\xxnote}[3]{}
\ifx\hidenotes\undefined
  \renewcommand{\xxnote}[3]{\color{#2}{#1: #3}}
\fi
\usepackage{amsthm}

\IEEEoverridecommandlockouts
\def\BibTeX{{\rm B\kern-.05em{\sc i\kern-.025em b}\kern-.08em
    T\kern-.1667em\lower.7ex\hbox{E}\kern-.125emX}}

\definecolor{darkgreen}{HTML}{006400}
\newcommand{\cyes}{\textcolor[HTML]{006400}{$\mathbf{\checkmark}$}}
\newcommand{\cno}{\textcolor{red}{$\mathbf{\times}$}}

\begin{document}


\title{\vspace{0.5cm}Don't Fool Me Twice: Adapting to Adversity in the Wild with Experience-Driven Reasoning
\thanks{
  \small{
  \emph{This work was performed when Navin Sriram was a visiting scholar at the Robotics Institute, Carnegie
  Mellon University.}}}
}

\author{
  Navin Sriram Ravie$^{1}$, \ Andrew Jong$^{2}$, \ Krrish Jain$^{2}$, \ John Liu$^{2}$, \ Omar Alama$^{2}$, \\
  Bijo Sebastian$^{\dagger 1}$, \ Sebastian Scherer$^{\dagger 2}$ \\
  \small{$^{1}$Department of Engineering Design, Indian Institute of Technology, Madras} \\
  \small{$^{2}$Robotics Institute, Carnegie Mellon University} \\
  \small{$\dagger$ Equal Advising}
}
\IEEEaftertitletext{
  \centering
  \includegraphics[width=\textwidth,height=0.3\textheight,keepaspectratio]{dfm2_main_figure_full_annotated.jpg}
  \captionof{figure}{Our framework \emph{Don't Fool Me Twice (DFM2)} enables robots to learn and adapt online to adversities in the wild. In the green region, the robot faces no adversity. However, in the red region it observes a perturbance in its trajectory. DFM2 reasons over semantics about the most likely cause (e.g. a fan) and learns to characterize its impact to the robot. When recognizing the same semantics later, the robot plans an appropriate path (blue) to circumvent the adversity.}
  \label{fig:abstract}
}
\maketitle

\begin{abstract}
In robotics, dangers and adversity modes are often embodiment-specific and relative to each agent. A frontier of autonomous mobile robotics is to enable agents to operate effectively in the wild in unseen unstructured environments, where it may not be possible to predetermine all possible dangers specific to a robot. Although recent work has used large foundation vision-language models (VLMs) to preemptively predict an exhaustive list of common-sense dangers, it remains difficult to capture possible interaction and embodiment-dependent adversities. We propose a continual learning framework for a mobile embodied agent to learn online from disturbances by using semantics to connect causes to anomalous effects, enabling better prediction and planning of the world in the future. Our framework, \emph{Don't Fool Me Twice}, first observes disturbances and narrates their effects on the robot; this description is augmented with visual context to query a VLM to infer possible causes; the local disturbance is characterized using kernel regression, which allows for efficient, few-shot modeling of transient anomalies. We leverage semantic voxel-centric modeling to estimate epistemic uncertainty, enabling richer downstream recovery by treating interaction-driven disturbances as learnable spatial behaviors. To test our framework, we explore four hypotheses and validate them with simulation and hardware.
\end{abstract}


\section{Introduction}
Autonomous robots deployed in open-world environments face embodiment-specific adversities that cannot be fully anticipated beforehand. These embodiment-specific adversities have effects that are often relative to each robot. For example, a wet floor poses a danger to a wheeled robot but not to a flying micro-aerial vehicle. Even between two given wheeled robots, the adversity may manifest differently due to differences in their tractive systems. As another example, mobile robots use different algorithms for localization and odometry estimation; each of these may fail under different edge cases: a Visual Odometry (VO) pipeline might suffer from overexposure or near textureless surfaces as we see with \cite{Labb__2018} \cite{orbslam}, and a LiDAR Inertial Odometry (LIO) system can fail near reflective surfaces \cite{nagata2026mirrordriftactuatedmirrorbasedattacks} \cite{damodaran}. Many mobile robots contains a system of complex algorithms for perception, planning and control, and their combination can compound unexpected failure modes. As the world moves towards more end-to-end learning based systems, it's imperative to also consider the edge cases where such trained models could fail.
\noindent While large vision-language models (VLMs) offer a shortcut \cite{pavone}\cite{sinha2024realtimeanomalydetectionreactive}\cite{oh2025languagecostproactivehazard}\cite{jongwiriyanurak2025vroastvisualroadassessment} by enumerating hazards from visual scenes, they lack awareness of embodiment-specific interactions, yielding over-conservative detections on things that pose no real threat to the specific robot.  This raises the key question: \textit{how can an embodied agent autonomously discover, characterize, and adapt to adversities through its own operational experience?}

We present \textbf{Don't Fool Me Twice (DFM2)}, a framework that builds a library of semantically associated adversities and their impacts through operational experience. We define an adversity as any disturbance that produces a measurable deviation in a robot's operational signals from nominal behavior.
By associating anomalous operational deviations with semantic features, DFM2 constructs a retrievable library of adversity experiences grounded in operational interaction. Each adversity is further characterized through semantic voxel-centric disturbance modeling, enabling the robot to predict the disturbance impact and adapt when similar hazards reappear. Our main contributions are:
\begin{itemize}
    \item \textbf{Experience-Driven Semantic Association:} We discover and characterize long-tail hazards by attributing 
anomalous operational signal deviations to dense visual semantics, building a personalized ``danger library'' through post-hoc reasoning.
    \item \textbf{Decoupled Few-Shot Interaction Modeling:} We constrain spatial geometry via semantic voxel priors and structured spatial disturbance models, reducing adaptation to a constrained optimization requiring only sparse interaction data.
    \item \textbf{Uncertainty-Aware Predictive Adaptation:} We estimate epistemic uncertainty via Bayesian Linear Regression over a fixed shape template, enabling proactive, conservative planning near novel hazards.
\end{itemize}

\begin{figure*}[t]
    \centering
    \includegraphics[width=0.9\linewidth]{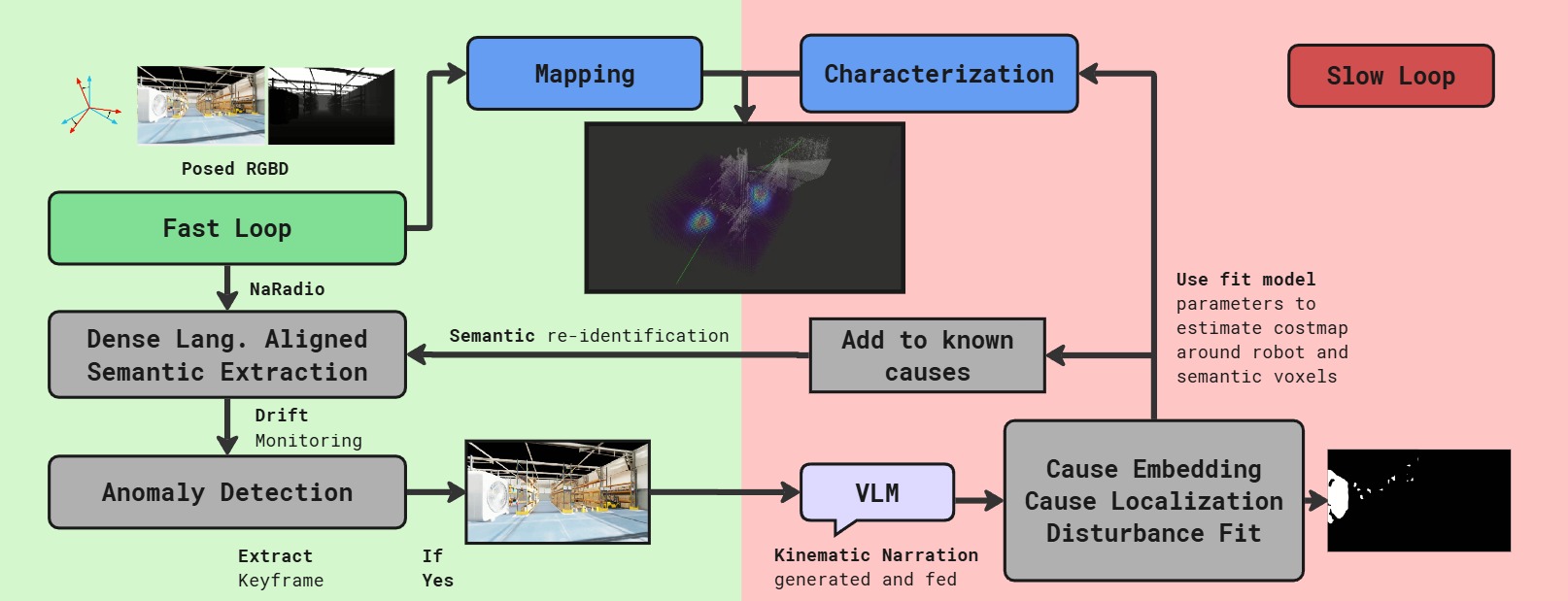}
    \caption{Proposed pipeline of \emph{Don't Fool Me Twice} showing its a) fast loop (green) and b) slow loop (red). As the robot
encounters an anomalous disturbance at (1), it stores and
reasons about the drift, finding out that it was the fan that was the source of disturbance. As it
resumes in the fast loop, it sees another fan at (2) and using the expanded danger library, it knows that it is an object of risk}
    \label{fig:method_fig}
\end{figure*}


\section{Related Work}

\subsection{Semantic Hazard Anticipation with VLMs}
Recent research has increasingly leveraged Vision--Language Models (VLMs) to provide semantic priors for proactive hazard assessment in unstructured environments. For example, \textit{Language as Cost (LaC)} \cite{oh2025languagecostproactivehazard} utilizes zero-shot VLM prompting to interpret visual scenes and map textual hazard cues into spatial costmaps via Gaussian cost propagation. Similarly, \textit{V-RoAst} \cite{jongwiriyanurak2025vroastvisualroadassessment} introduces a VLM-based visual question-answering (VQA) framework to classify infrastructure safety attributes at regular road segments. Chakraborty \textit{et al.} \cite{chakraborty2026autonomouslaboratorysafetymonitoring} further addressed the perceptual limitations of VLMs in raw image recognition by proposing a ``scene-graph--guided alignment" that structures environmental cues into relational object graphs. While these efforts demonstrate that foundation models encode rich hazard concepts, they predominantly rely on a proactive ``query at regular intervals" strategy. This approach typically casts a wide net by cataloging all potential adversities based on internet-scale common-sense priors, which introduces conservatism, often forcing detours around objects that look dangerous but pose no actual threat to a specific robot.

\subsection{Few-Shot Interaction Modeling }

Modeling robot–environment interactions  has traditionally been studied through online system identification \cite{ljung1998system}, which assumes a fixed parametric structure and sufficient training points, assumptions that often break down during transient events. To relax these constraints, kernel-based regression and Gaussian Processes (GPs) have been widely used to model unknown dynamics and disturbance fields \cite{gaussianml, deisenroth2015distributedgaussianprocesses}.
Related approaches learn residual dynamics or disturbance fields on top of nominal models \cite{Kocijan01122005} or estimate spatial fields using basis functions or grids \cite{Faessler_2018, OCallaghan2012GaussianPO}, but typically treat residuals as unstructured functions or require dense exploration. Meta-learning and fast adaptation methods \cite{finn2017maml, nagabandi2019learning} mitigate data scarcity via task-level pre-training, yet offer limited robustness to genuinely novel, long-tail hazards.

\subsection{Continual Learning and Runtime Monitoring}

 Modern frameworks have moved toward hierarchical monitoring loops and retrieval-augmented memory to prevent catastrophic forgetting. Recent developments  have solidified the utility of slow-fast loops. The Monitor-Analyse-Plan-Execute-Knowledge (\textit{MAPE-K}) architecture has evolved into variants like \textit{MAPLE-K}. \textit{MAPLE-K} enhances this structure by embedding a continuous learning (L) step, which integrates semantic grounding in our case. Our framework, DFM2, synthesizes strengths from several contemporary systems: \textbf{AESOP}\cite{sinha2024realtimeanomalydetectionreactive}: Provides high-rate anomaly detection via embedding similarity but lacks semantic causality. \textbf{FORTRESS}\cite{pavone}: Uses VLMs to identify safe fallback goals but suffers from over-conservativeness due to binary hazard labels. \textbf{DRAE}\cite{long2025draedynamicretrievalaugmentedexpert}: Utilizes a non-parametric Bayesian memory to store specialized ``experts" without corruption.


\section{Preliminaries}
\subsection{System Definitions}
We consider a mobile robot operating in a 3D environment tasked with following a reference trajectory $\tau_{\text{ref}}$ generated by a high-level planner. We assume the robot is equipped with a \textit{reliable low-level controller} 
that maps control inputs $\mathbf{u} \in \mathcal{U}$ to velocities, maintaining a bounded tracking error in nominal, disturbance-free conditions. We define an \textit{adversity} as any disturbance that degrades a robot's ability to execute its intended task, manifesting as \textit{measurable deviations in operational signals}. To capture these varied failure modes across heterogeneous robotic platforms and sensory configurations, we introduce a generalized, time-varying operational monitoring signal $\nu(t)$. In this work, the signal $\nu(t)$ is dynamically mapped based on the active embodiment and context:
\begin{equation}
\nu(t) = 
\begin{cases} 
\lVert \mathbf{x}_{\text{obs}}(t) - \mathbf{x}_{\text{ref}}(t) \rVert, & \text{(Trajectory Tracking Error)} \\
\text{Tr}\left(\mathbf{\Sigma}_{\text{pose}}(t)\right), & \text{(State Estimation Uncertainty)}
\end{cases}
\end{equation}
where $\mathbf{x}_{\text{ref}}(t)$ and $\mathbf{x}_{\text{obs}}(t)$ represent the reference and executed coordinates respectively, and $\text{Tr}\left(\mathbf{\Sigma}_{\text{pose}}(t)\right)$ denotes the trace of the state estimator's pose covariance matrix. 
An adversity event is then parameterized via a binary operational indicator function $a(t) \in \{0, 1\}$ defined as:
$a(t) = \mathbb{I}\left(\nu(t) > \theta\right)$
where $\mathbb{I}(\cdot)$ is the indicator function and $\theta$ is a context-specific detection threshold. To precisely calibrate $\theta$ and characterize baseline sensor and actuator noise floor variations, the robot executes a prior calibration run path within a nominal, anomaly-free space.

\subsection{Problem Statement}
Given an active reference trajectory $\tau_{\text{ref}}$ and a continuous stream of online operational measurements $\nu(t)$, the objective of this work is to build an experience-driven reasoning architecture that can:
\begin{enumerate}
    \item \textbf{Detect} anomalous operational adversities in real-time when the indicator function turns active ($a(t) = 1$).
    \item \textbf{Infer} the latent semantic cause $c \in \mathcal{C}$ of the localized disturbance post-hoc by leveraging a Vision-Language Model (VLM).
    \item \textbf{Learn} a highly sample-efficient predictive model $f: \mathbb{R}^3 \to \mathbb{R}$ of the localized disturbance field magnitude alongside its epistemic uncertainty.
\end{enumerate}



\section{Methodology}

After calibration, the robot follows its planned path $\tau_{\text{ref}}$ while continuously monitoring for any perceived disturbances. The architecture is organized into two complementary pathways: a \textit{fast loop}  and a \textit{slow loop}.

\subsection{3D Semantic Search for Adversity}

At the core of the fast loop is a high-frequency perception module that detects risk-relevant objects using the dynamic danger library $\Omega_t$, which stores embeddings of previously observed adversity causes. Detection is performed via embedding similarity $\texttt{sim}(\cdot,\cdot)$ search between encoded image features and the danger library, offering spatial granularity and richer generalizability over text-based matching. We build on the NaRadio encoder ($\texttt{Embed}_{vision}$) from RayFronts\cite{rayfronts} with OpenVDB \cite{ovdb} based mapping backbone. NaRadio leverages the RADIO vision foundation model encoder\cite{radio} which distills features from diverse vision encoders (SAM for segmentation, DINOv2 for context, and SigLIP\cite{siglip} for language alignment). Incoming RGB images are thus transformed by NaRadio into spatially consistent feature maps at the pixel level in the SigLIP space ($\texttt{Embed}_{lang}$) in the form of dense language aligned features. By performing $\texttt{sim}(\cdot,\cdot)$, we extract hotspot masks through simple thresholding. These pixels are projected, voxelized and integrated into the underlying OpenVDB generated voxel map. 


\subsection{Anomaly Detection and Reasoning}
\begin{figure}[t]
    \centering
    \includegraphics[width=1\linewidth]{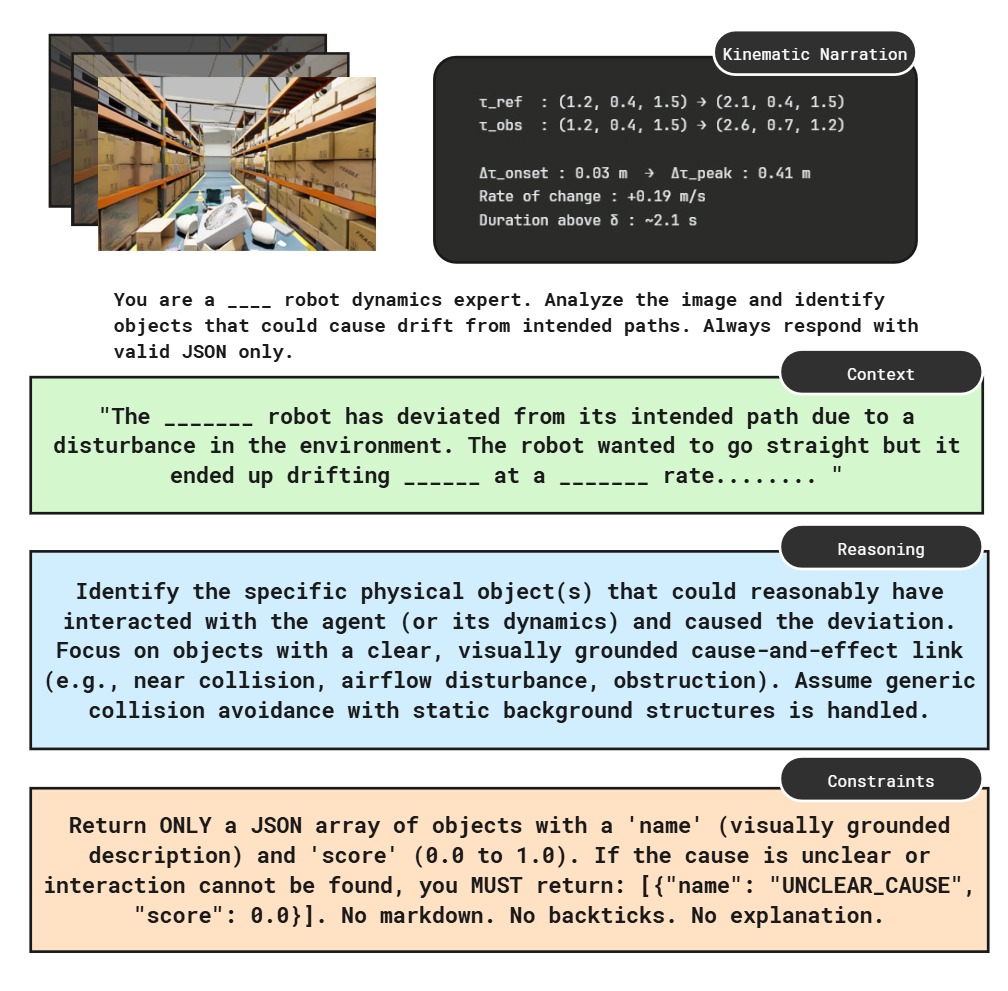}
    \caption{ Narration workflow showcasing the event driven narration framework showing each aspect of the prompt}
    \label{fig:narration}
\end{figure}

 The system continuously evaluates trajectory tracking error  against the calibrated threshold. When $a(t) = 1$, an anomaly event is flagged. The anomaly detection module records the onset and offset of each breach and buffers multimodal data (RGB, depth, pose throughout the breach event). Upon anomaly detection, the system transitions into a narrative construction and reasoning stage. It synthesizes the intended ($\tau_{\text{ref}}$) and actual ($\tau_{\text{obs}}$) behaviors and analyzes relevant metrics (e.g., $\nu(t)$, rate of change, curvature) as shown in Fig.~\ref{fig:narration}. Instead of querying the VLM blindly we provide a multimodal grounding. The system generates a multimodal experience log, similar to \cite{pavone} but with a linguistically-grounded narrative about the experience. Anomaly detection acts as a better event-driven key-frame extraction method to feed the VLM compared to most methods that simply query a VLM at regular intervals. Event-driven querying reduces conservatism, filters candidates by spatial proximity, and isolates the single causal object, directly reducing false positives. We test our narration framework with different models and techniques.

\subsection{Cause Localization} The VLM outputs a textual description of the inferred disturbance cause, for example, \emph{``fan"}. We encode this textual description into its embedding set \(\mathcal{E}_t = \texttt{Embed}_{lang}(\text{vlm-description})\) and extract dense per-pixel embeddings \(\mathcal{E}_p = \texttt{Embed}_{vision}(\text{image})\) from the corresponding image. We compute similarity scores \(s_p = \texttt{sim}(\mathcal{E}_t, \mathcal{E}_p)\) between the VLM description's and each pixel's embedding from $\mathcal{E}_p$. Using these similarity scores as weights, we compute a weighted average over all pixel embeddings to obtain a richer, contextually grounded embedding.

\vspace{5pt}

\begin{equation}
    s_p = \texttt{sim}(\mathcal{E}_t, \mathcal{E}_p), \quad \quad \mathbf{e} = \frac{\sum_p s_p \mathcal{E}_p}{\sum_p s_p}.
\end{equation}

Upon estimating this embedding, we use it to localize the cause object of the anomaly corresponding to the narration as shown in Fig.~\ref{fig:characterization}. This embedding is stored and then used for re-identification.

\subsection{Semantic Adversity Association and Characterization}

We model the predictive effect of a hazard on the agent by learning how different adversities manifest in downstream tracking error (or other adversity signals) as discussed before. Kernel regression is a natural choice for characterizing these behaviours \cite{siva}. 
\subsubsection{Voxel-Centric Kernel Regression}
\begin{figure}[t]
    \centering
    \includegraphics[width=1
    \linewidth]{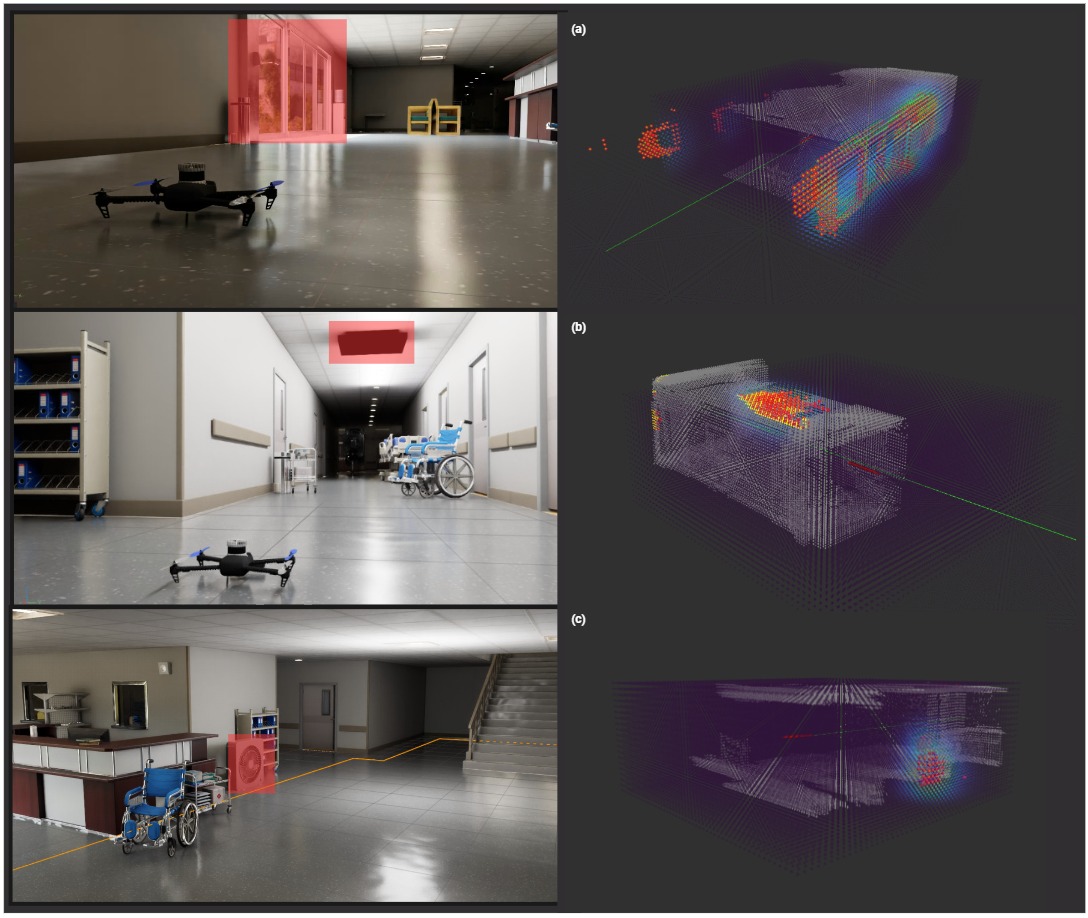}
    \caption{Map with the voxel centric characterization in space for (a) Window (b) Vent and a (c) Fan causing a mobile drone robot to drift. Red corresponds to hazard voxels, disturbance field is estimated around these unsafe voxels}
    \label{fig:characterization}
\end{figure}

We model adversity at the voxel level, where each semantic voxel encodes the local effect on the robot's pose. Let $\mathcal{V}$ denote the set of semantic voxels associated with a hazard, with centers $c_j \in \mathbb{R}^3$. We treat adversities as transferable behaviors by enforcing identical kernels across voxels, using anisotropic RBF kernels whose locality confines each voxel's influence spatially while capturing directional robot-environment interactions. For a query position $x \in \mathbb{R}^3$, the disturbance is
\begin{equation}
    f(x) = A \sum_{j \in \mathcal{V}} 
    \exp\!\left(-\tfrac{1}{2} Q_\ell(x - c_j)\right) + b,
\end{equation}
with anisotropic quadratic form
\begin{equation}
    Q_\ell(x - c_j) = 
    \frac{(x - c_j)_x^2 + (x - c_j)_y^2}{\ell_{xy}^2}
    + \frac{(x - c_j)_z^2}{\ell_{z}^2},
\end{equation}
where $\ell_{xy}, \ell_z > 0$ govern horizontal and vertical disturbance spread. Given trajectory tracking errors $d(x_i)$ along the planned path, we fit $(A, b, \ell_{xy}, \ell_z)$ by minimizing
\begin{equation}
\min_{\ell_{xy}, \ell_z} 
    \frac{1}{N} \sum_{i=1}^N 
    \big(d(x_i) - f(x_i)\big)^2.
\end{equation}
This yields fine-grained, geometry-aware fields that generalize across hazards of varying size and pose (e.g., large vs. small fans) while remaining computationally efficient. Parameters $(\mathcal{V}, A, b, \ell_{xy}, \ell_z)$ learned for one hazard are stored in the danger library with the associated cause; example characterizations are shown in Fig.~\ref{fig:characterization}.

\subsubsection{Epistemic Uncertainty Estimation}
The kernel regression above gives a deterministic mean estimate; for safety we additionally quantify \textit{epistemic uncertainty} — uncertainty due to lack of data. We keep this tractable via our semantic prior: the disturbance's \textit{spatial shape} ($\ell_{xy}, \ell_z$) is fixed by object class, while its \textit{intensity} ($A, b$) is treated as uncertain, reducing the problem to Bayesian Linear Regression over a fixed basis. With feature vector
\begin{equation}
\bm{\phi}(\mathbf{x}) = 
\begin{bmatrix}
\phi_{\text{ker}}(\mathbf{x}) \\
1
\end{bmatrix}, \quad
\phi_{\text{ker}}(\mathbf{x}) = 
\sum_{j \in V} 
\exp\left(
-\frac{1}{2} Q_{\ell}(\mathbf{x}-\mathbf{c}_j)
\right),
\end{equation}
and weights $\mathbf{w} = [A, b]^\top$, the posterior covariance is
\begin{equation}
    \bm{\Sigma}_w = \sigma_n^2 (\bm{\Phi}^\top \bm{\Phi})^{-1},
\end{equation}
where $\bm{\Phi}$ is the design matrix of collected observations and $\sigma_n^2$ is the aleatoric noise variance, estimated online from residuals. The predictive variance at a query state $\mathbf{x}_*$ is
\begin{equation}
    \sigma^2(\mathbf{x}_*) = \underbrace{\sigma_n^2}_{\text{Aleatoric}} + \underbrace{\bm{\phi}(\mathbf{x}_*)^\top \bm{\Sigma}_w \bm{\phi}(\mathbf{x}_*)}_{\text{Epistemic}}.
\end{equation}
Where the shape template predicts strong disturbance but data is scarce, uncertainty in $A$ is amplified; where the template is zero, uncertainty collapses to the ambient noise level, preventing false positives.


\section{Evaluation}

\subsection{Hypotheses}

\textbf{Hypothesis 1:} \emph{An online experience-driven disturbance library can capture edge-case disturbances that may be otherwise missed, leading to higher survival rate.}

\textbf{Hypothesis 2:} \emph{An event-driven VLM querying mechanism mitigates over-conservativeness of preemptive libraries, leading to faster arrival times and shorter path lengths.}

\textbf{Hypothesis 3:} \emph{Characterizing the disturbance's spatial impact, rather than just treating semantics as binary obstacles, leads to more effective downstream planning.}

\textbf{Hypothesis 4:} \emph{DFM2's flexible definition of an adversity signal demonstrates applicability across qualitatively distinct failure modes and across multiple robot embodiments and navigation 
stacks.}

\subsection{Baselines}
We compare against 3 baselines that ablate key design choices: semantic grounding, library construction strategy (preemptive vs.\ experience-driven), and characterization.

\textbf{DROAN-GL (Pure Geometric):} A pure geometric planner that performs obstacle avoidance using voxel-projected occupancy from depth sensors, with no semantic querying or disturbance characterization. This is a GPU accelerated version based on \cite{droan}. All baselines leverage DROAN-GL for path tracking and obstacle avoidance with the trajectory-selection cost function modified with a simple weighted sum formulation proportional to the disturbance magnitude.

\textbf{Pro-Active Hazard Reasoning:} Following LaC~\cite{oh2025languagecostproactivehazard} and Ganai et al.~\cite{pavone}, we implement a preemptive baseline using NaRadio-based language-embedding similarity lifted to a 3D navigation costmap. We also include \textbf{Pro-Active-Avoid}, which queries the VLM for required clearance and constructs spherical avoidance regions around grounded semantics.

\subsection{Ablations}
\begin{figure}[t]
    \centering
    \includegraphics[width=1\linewidth]{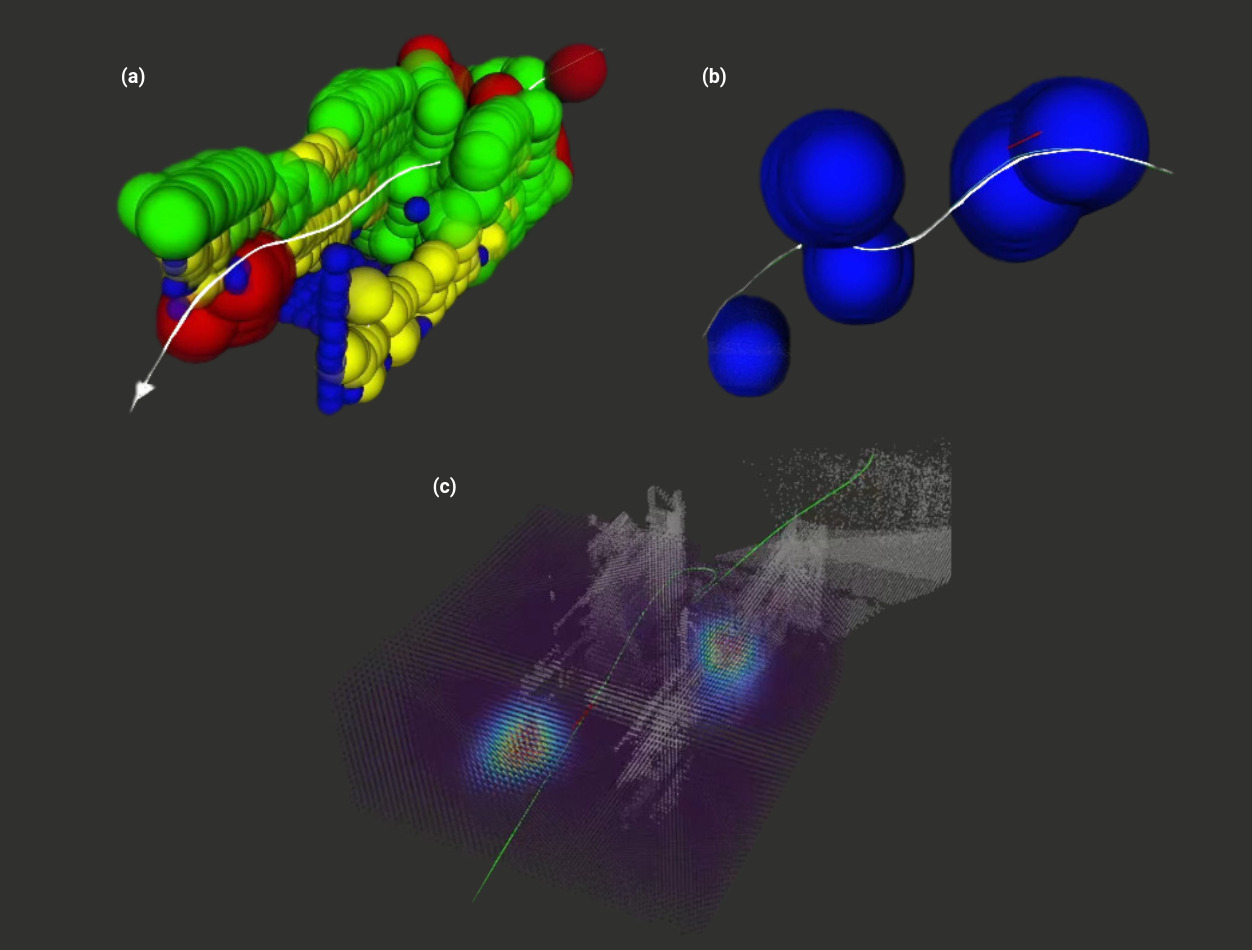}
    \caption{(a) Preemptive Hazard Anticipation, each colour is a distinct semantic label deemed to be unsafe with the sphere radii reflecting the clearance around it (b) DFM2-Avoid with the avoidance regions around the unsafe semantics (c) DFM2 showcasing the characterization and the executed path }
    \label{fig:traj_viz}
\end{figure}
\textbf{DFM2-Avoid Fixed Radius:} Uses the experience-driven danger library but replaces disturbance characterization with a fixed spherical avoidance radius. Tested at 1.5m (DFM2-Avoid-1.5) and 2.5m (DFM2-Avoid-2.5) to isolate the contribution of adaptive spatial modeling (H3).

\textbf{DFM2 (Full Method):} Combines VLM-based grounding, experience-driven library construction, and voxel-centric disturbance characterization.

\begin{table*}[th]
\centering
\caption{Comparison of baselines and ablations. Each row systematically isolates different components to validate our core hypotheses.}
\label{tab:baselines}
\begin{tabular}{@{}lccccc@{}}
\toprule
\textbf{Method} & \textbf{3D Proj.} & \textbf{Semantic} & \textbf{Semantic} & \textbf{Disturbance} & \textbf{Planner} \\
 & & \textbf{Grounding} & \textbf{Querying} & \textbf{Characterization} & \\
\midrule
DROAN-GL & \cyes & Voxel proj. & \textcolor{red}{None} & \cno & Traj. Library \\
Pro-Active Hazard Reasoning & \cyes & Voxel proj. & \textcolor{orange}{Pre-emptive} & \cyes & Traj Library \\
Pro-Active-Avoid & \cyes & Voxel proj. & \textcolor{orange}{Pre-emptive} & \cno & Traj Library \\
DFM2-Avoid Fixed Radius=2.5 & \cyes & Voxel proj. & \textcolor{darkgreen}{Exp.-driven} & \cno & Traj. Library \\
DFM2-Avoid Fixed Radius=1.5 & \cyes & Voxel proj. & \textcolor{darkgreen}{Exp.-driven} & \cno & Traj. Library \\
\textbf{DFM2 (Ours)} & \cyes & Voxel proj. & \textcolor{darkgreen}{Exp.-driven} & \cyes & Traj. Library \\
\bottomrule
\end{tabular}
\end{table*}


\section{Simulation Experiments}
We implement our simulation using NVIDIA Isaac Sim~\cite{NVIDIA_Isaac_Sim}, which provides realistic physics and photorealism.\newline

\noindent\textbf{Robot Dynamics, Control, and Perception: }
We simulate a flying quadrotor. To simulate the dynamics of a quadrotor robot, we use the Isaac Pegasus extension \cite{jacinto2024pegasus}. Pegasus simulates forces and torques from each propeller to drive the body mass while outputting sensor data to PX4 \cite{meier2015px4}. For quadrotor control, we implement our own software stack, \emph{AirStack}, that sends velocity and yaw control commands ($v_x, v_y, v_z, \omega$) to the PX4 controller.
The robot uses RGB-D camera sensors (360$\times$480). We simulate physical disturbances (fans, vents, windows) using Isaac Sim's force-field APIs. Experiments run in a modular warehouse (IsaacSim assets) with seven aisles and an open area. We spawn 2--3 disturbance sources per aisle with randomized orientations. The drone must traverse an aisle to reach a goal, with racks of dense semantic clutter flanking each aisle to challenge causal attribution. We run 25 randomized navigation trials each lasting $>$30s.
\subsection{Cause Reasoning}
We evaluate event-driven causal attribution against traditional anticipative querying (every 4s). We curated a dataset of aerial and wheeled robotic failures across cluttered and sparse environments. In a 40s episode, anticipative querying generates 30--40 potential hazards; with a singular root cause, this yields precision $<5\%$. In contrast, the DFM2 mechanism isolates the disturbance event before querying the VLM. Among the models tested in Table \ref{tab:vlm_comparison}, o4-mini (CoT) achieved the highest performance (0.96 MRR), demonstrating that ``Chain-of-Thought" reasoning is beneficial for pruning irrelevant visual artifacts. Notably, Gemini 2.5 Flash emerged as the most efficient non-reasoning model, providing a precision of $82.9\%$ while maintaining a low output count $(O_p=1.9)$. These results suggest that current VLMs can provide reliable semantic attribution when the search space is constrained to a localized temporal failure window.

\begin{table}[t]
\centering
\caption{Causal attribution across VLMs: DFM2 post-hoc mechanism vs.\ anticipative querying.}
\label{tab:vlm_comparison}

\resizebox{\columnwidth}{!}{%
\begin{tabular}{@{}llcccc@{}}
\toprule
\textbf{Mechanism} & \textbf{Model} & \textbf{Prec.} ($\%$) ($\uparrow$) & \textbf{MRR} ($\uparrow$) & \textbf{O_p} ($\downarrow$) \\
\midrule

\multirow{5}{*}{\shortstack[l]{Post-hoc \\ (DFM2)}} 
 & GPT-4o-mini   & 75.0 & 0.94 & 2.7\\
 & o4-mini (CoT)      & 85.7 & 0.96 & 1.6  \\
 & Sonnet 4.5    & 65.7 & 0.88 & 2.4  \\
 & Gemini 2.5 F. & 82.9 & 0.94 & 1.9 \\

\bottomrule
\end{tabular}%
}
\end{table}







\subsection{Results}
 We show the quantitative results of randomized control trials in Table~\ref{tab:sim_quant_results_2}. \textsc{DFM2} consistently outperforms the baselines in survival rate and cumulative disturbance, validating H1--H3. \textsc{DROAN-GL} crashes most frequently without semantic awareness; preemptive methods improve safety but suffer over-conservatism (H2). Against binary ablations (\textsc{DFM2-Avoid-1.5/2.5}), \textsc{DFM2}'s continuous disturbance fields (H3) enable emergent behaviors like stopping, reversing, and rerouting around hazards rather than blind persistence, as shown in Fig.~\ref{fig:traj_viz}.

\begin{table*}[th]
\centering
\caption{Baselines and Ablations on \textbf{successful trials}. Bold = best. $\uparrow$ higher is better, $\downarrow$ lower is better. Over 25 trials.}
\label{tab:sim_quant_results_2}
\resizebox{\textwidth}{!}{%
\begin{tabular}{@{}lcccccccc@{}}
\toprule
\textbf{Method} & 
\makecell{\textbf{Arrival Time} \\ $T_{\text{arr}}$ (s) $\downarrow$} & 
\makecell{\textbf{Path Length} \\ $L_{\text{path}}$ (m) $\downarrow$} & 
\makecell{\textbf{Survival} \\ $R_{\text{survive}}$ (\%) $\uparrow$} &  
\makecell{\textbf{Cum.} \\ \textbf{Disturbance} (m) $\downarrow$} & 
\makecell{\textbf{Norm. Cum.} \\ \textbf{Disturbance} (cm) $\downarrow$} \\
\midrule
DROAN-GL & 30.2 $\pm$ 3.89 & 31.7 $\pm$ 8.7 & 40.9 &  10.3 $\pm$ 9.4 & 31.4 $\pm$ 28.5  \\
Pro-Active & 31.4 $\pm$ 3.33 & 34.0 $\pm$ 10.9 & 59.1  & 8.0 $\pm$ 6.92 & 22.06 $\pm$ 18.25 \\
Pro-Active-Avoid & 31.5 $\pm$ 3.30 & 33.4 $\pm$ 11.4 & 54.5  & 10.2 $\pm$ 10.84 & 24.3 $\pm$ 22.2 \\
DFM2-Avoid-1.5 & 33.2 $\pm$ 4.56 & 36.4 $\pm$ 11.8 & 68.2 & 12.9 $\pm$ 12.2 & 11.5 $\pm$ 36.15 \\
DFM2-Avoid-2.5 & 33.9 $\pm$ 4.83 & 33.2 $\pm$ 9.6 & 72.7 & 7.4 $\pm$ 12.5 & 18.8 $\pm$ 30.63 \\
\midrule
\textbf{DFM2 (Ours)} & \textbf{31.04 $\pm$ 3.68} & \textbf{29.3 $\pm$ 6.5} & \textbf{81.8} & \textbf{2.9 $\pm$ 4.8} & \textbf{9.2 $\pm$ 13.68}  \\
\bottomrule
\end{tabular}%
}
\end{table*}

\section{Hardware Experiments}
The simulation experiments validate DFM2 against external physical disturbances. We now evaluate a different kind of adversity: perception-induced degradation, in which the robot's internal state estimator is compromised by environmental visual properties quite literally ``fooling" the robot into believing it has moved. We validate Hypothesis 4 (H4) by deploying DFM2 on a wheeled mobile robot under two such failure modes. In contrast to wind fields, the disturbances studied here originate from sensor degradation.

 \noindent\textbf{Robot Dynamics, Control, and Perception: }To evaluate this setting, we deploy DFM2 on a wheeled mobile robot equipped with an Intel RealSense D455 RGBD camera, a Velodyne VLP32 LiDAR, and an Xsens IMU. The platform performs local navigation using the Dynamic Window Approach (DWA) with laser scan based collision avoidance augmented with DFM2-derived disturbance costs, used through a simple artificial potential field formulation for adaptation. We evaluate DFM2 using the RTAB-Map RGBD-inertial odometry pipeline \cite{Labb__2018}, which combines feature-based visual correspondence estimation, RGBD geometric constraints, IMU-assisted motion prediction, and local graph optimization for pose tracking. We study two commonly observed but often under-characterized perception failures:
\begin{figure}[t]
    \centering
    \includegraphics[width=0.75\linewidth]{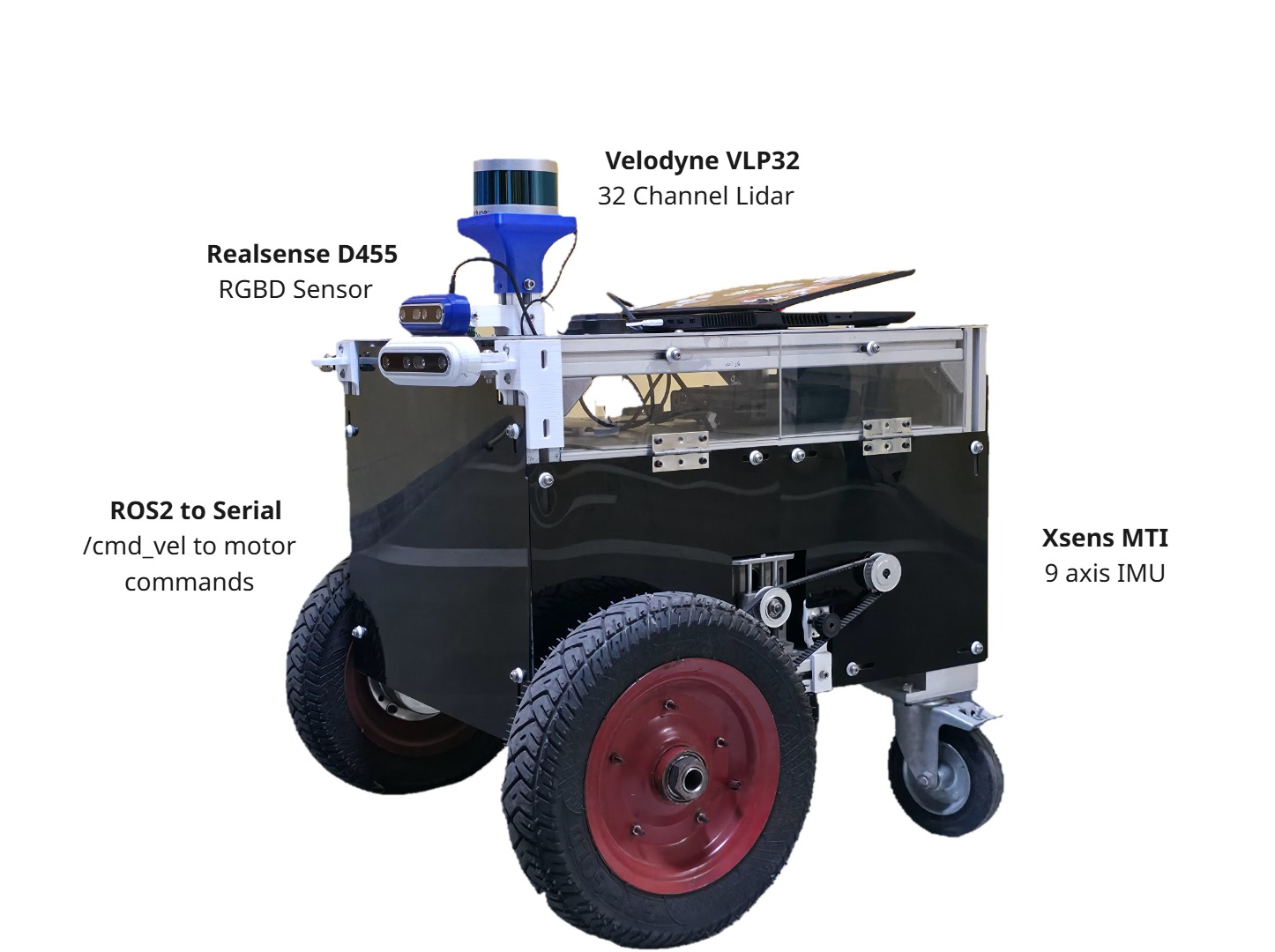}
    \caption{Differential drive wheeled mobile robot platform running the complete stack on a commercial laptop}
    \label{fig:hard_placeholder}
\end{figure}

\begin{itemize}

\item \textbf{Textureless Surface Degeneration:} Large textureless surfaces reduce the number of reliable visual correspondences available to visual odometry systems, producing unstable tracking and increased estimator uncertainty. We induce this failure mode using textureless and visually uniform bedsheets positioned near turning regions in narrow corridors, where localization behavior depends strongly on viewpoint and field-of-view occupancy rather than obstacle distance alone.






\begin{figure}[t]
    \centering
    \includegraphics[width=1.0\linewidth]{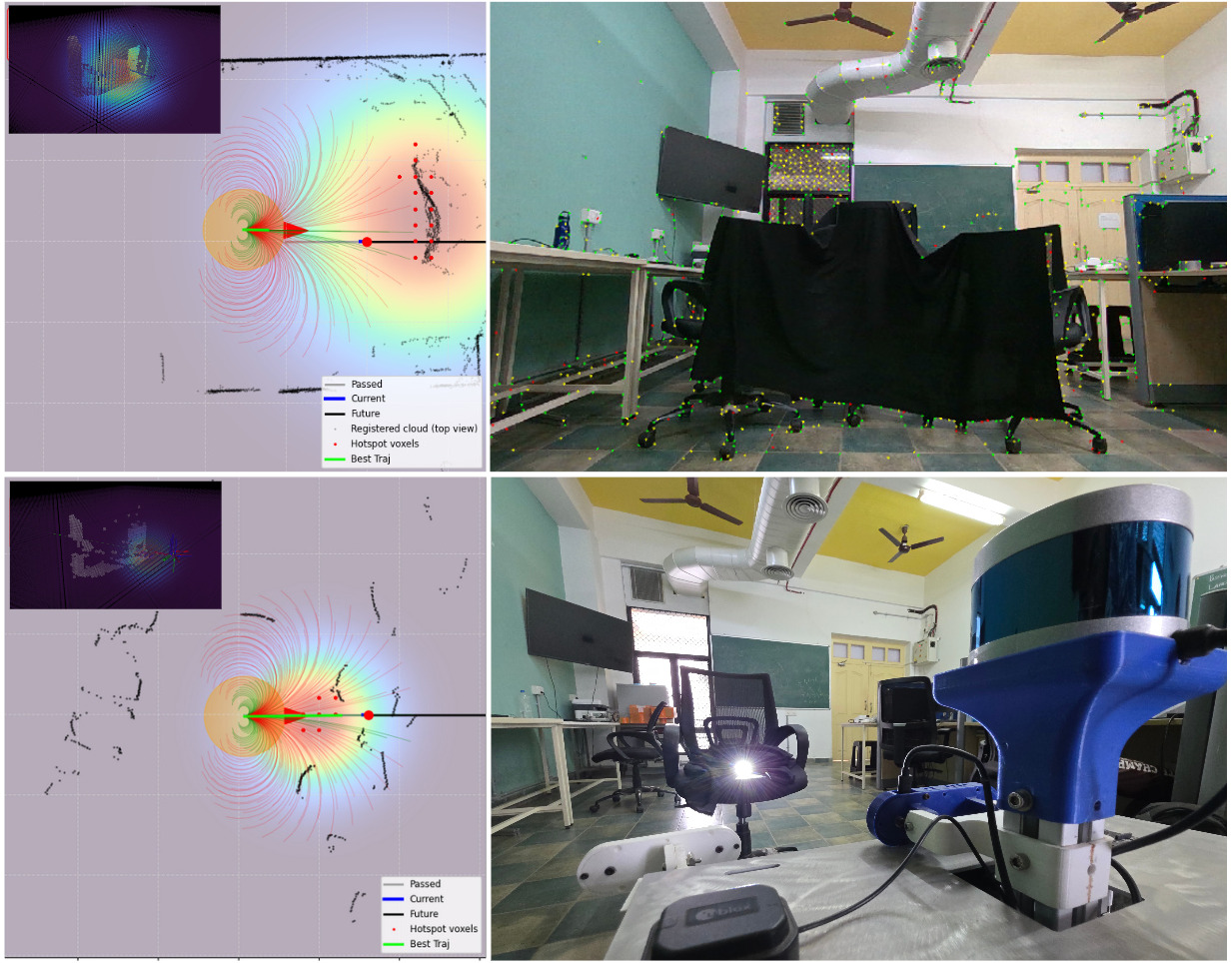}
    \caption{Top: Featureless-surface demonstration (black sheet). (a) 3D DFM2 characterization (red→violet: decreasing intensity) with 2D occupancy grid and DWA primitives overlaid; (b) keypoint extraction showing matches everywhere except the black surface. Bottom: Overexposure demonstration. (a) 3D DFM2 characterization with occupancy grid and DWA primitives overlaid; (c) bright pocket torch facing head-on.}
    \label{fig:sheet_placeholder}
\end{figure}
\item \textbf{Infrared Sensor Overexposure:} Active infrared depth sensors are susceptible to strong illumination and IR interference, resulting in degraded depth observations and unstable odometry estimation. We reproduce this phenomenon by exposing the RealSense D455 to illumination from pocket torches placed in the environment.
\end{itemize}

We monitor the trace of the pose covariance matrix $\text{Tr}(\Sigma)$ as the adversity signal and the disturbance model $f(\mathbf{x})$ characterizes the spatial distribution of this covariance inflation. We compare DFM2 exclusively against DWA because alternative baselines fail within the narrow 2.5m corridor (robot width: 0.4m). Other avoidance mechanisms yield infeasible trajectories, while Pro-Active-Avoid constructs large spherical avoidance regions (0.75–2.5m radius) that make navigation geometrically impossible. Resolving this would require environment-specific manual tuning, which defeats the purpose of this work. All onboard computation was performed on a consumer-grade laptop equipped with a Ryzen 7 5600X CPU, 16GB RAM, and an RTX 3060 GPU, running DFM2 and the control stack with all the sensors.

\subsection{Evaluation Metrics}
Unlike the simulation experiments, perception-driven failures in real-world systems often emerge before significant trajectory drift becomes externally observable. We therefore evaluate localization health using internal estimator consistency metrics in addition to qualitative navigation behavior.

\begin{itemize}

\item \textbf{Inlier Ratio} $(R_{inlier})$:  
Measures the fraction of geometrically consistent feature correspondences used by the odometry backend:
\begin{equation}
R_{inlier} = \frac{N_{inlier}}{N_{matches}}
\end{equation}
Lower values indicate degraded visual observability and unstable tracking.

\item \textbf{Pose Covariance Trace} $(\mathrm{Tr}(\Sigma))$:  
Measures estimator uncertainty using the trace of the pose covariance matrix reported by the odometry system.
Higher covariance indicates reduced estimator confidence.

\item \textbf{Degradation Duration} $(T_{deg})$:  
Measures cumulative time spent in degraded localization states, determined using covariance thresholding.
\end{itemize}

\subsection{Results}
The results in Table \ref{tab:hardware_quant_results} demonstrate that DFM2 improves localization robustness under perception degradation caused by large textureless regions validating H4. H1–H3 establish that DFM2 works; H4 establishes why it works across contexts. A wheeled robot experiencing localization drift from a textureless surface and a quadrotor experiencing trajectory deviation from a wind field are, from DFM2's perspective, identical events. Compared to baseline DWA navigation, DFM2 maintains a substantially higher feature correspondence inlier ratio while simultaneously reducing pose covariance inflation and cumulative localization failure duration. Qualitatively, the baseline planner frequently approaches feature-deprived regions too aggressively during turning maneuvers, resulting in unstable tracking and intermittent odometry degradation. In contrast, DFM2 adapts robot motion to reduce exposure to perceptually degraded viewpoints, indirectly improving visual observability and estimator stability before catastrophic failure occurs.

\begin{table}[t]
\centering
\caption{Hardware evaluation under perception degradation caused by large textureless surfaces during turning maneuvers. Bold = best. $\uparrow$ higher is better, $\downarrow$ lower is better. Averaged over 40 successful trials.}
\label{tab:hardware_quant_results}

\small
\begin{tabular}{@{}lccc@{}}
\toprule

\textbf{Method} &

\makecell{\textbf{Inlier} \\
\textbf{Ratio} $\uparrow$} &

\makecell{\textbf{Cov.} \\
\textbf{Trace} $\downarrow$} &

\makecell{\textbf{Degradation} \\
\textbf{Time (s)} $\downarrow$} \\

\midrule

DWA
& $0.28 \pm 0.23$
& $2.60 \pm 2.23$
& $8.33 \pm 7.62$ \\

\textbf{DWA + DFM2}
& $\mathbf{0.45 \pm 0.20}$
& $\mathbf{1.04 \pm 1.75}$
& $\mathbf{3.70 \pm 6.66}$ \\

\bottomrule
\end{tabular}
\end{table}

\section{Conclusion and Future Work}

We present Don't Fool Me Twice (DFM2), a framework for continual, experience-driven adaptation to embodiment-specific adversities in unstructured environments. By coupling event-driven kinematic narration with semantic voxel-centric kernel regression, DFM2 attributes online disturbances to causal semantics, characterizes their spatial extent, and proactively plans around them upon re-encounter. Across 25 randomized simulation trials, DFM2 achieves an 81.8\% survival rate and reduces cumulative disturbance by more than 3x relative to the next best method, validating all four core hypotheses.
 Our framework has been validated across diverse adversity modes and embodiments. A primary limitation of DFM2 is its reliance on visual observability of the failure cause; indirect indicators (e.g., a ``wet floor" sign rather than the puddle itself) can sometimes substitute, but this remains constraining. Attribution also depends on a single VLM call per encounter, making it brittle in cluttered scenes; accumulating evidence across multiple interactions, verifying causality, and leveraging fallback or ensemble reasoning are natural next steps as multimodal models improve. Finally, DFM2's disjoint, engineered modules may compound errors at their boundaries; an end-to-end differentiable approach tightly coupling perception, causal reasoning, and disturbance modeling could address this.

\section{Acknowledgements}
We thank the AirLab at Carnegie Mellon University and the Autonomous Systems Lab for their contributions to the development of this framework. We also thank John Dolan and Rachel Burcin for facilitating the RISS program and for their continued support. We are grateful for insightful discussions with Jay Patrikar, Jay Karhade, Andrea Bajcsy, and the Intent Lab. Finally, we thank Sayooj Raveendran, Satyam Singh, Dipshikha Hazari, John Keller, Seungjae Baek, and AirLab’s small drone team for their assistance with system deployment. This work at Carnegie Mellon University was supported by the Defense Science and Technology Agency (DSTA) under contract \texttt{\#DST000EC124000205}.

\bibliographystyle{IEEEtran}
\bibliography{references} 

@misc{pavone,
  title={Real-Time Out-of-Distribution Failure Prevention via Multi-Modal Reasoning}, 
  author={Milan Ganai and Rohan Sinha and Christopher Agia and Daniel Morton and Marco Pavone},
  year={2025},
  eprint={2505.10547},
  archivePrefix={arXiv},
  primaryClass={cs.RO},
  url={https://arxiv.org/abs/2505.10547}
}

@article{orbslam,
   title={ORB-SLAM3: An Accurate Open-Source Library for Visual, Visual–Inertial, and Multimap SLAM},
   volume={37},
   ISSN={1941-0468},
   url={http://dx.doi.org/10.1109/TRO.2021.3075644},
   DOI={10.1109/tro.2021.3075644},
   number={6},
   journal={IEEE Transactions on Robotics},
   publisher={Institute of Electrical and Electronics Engineers (IEEE)},
   author={Campos, Carlos and Elvira, Richard and Rodriguez, Juan J. Gomez and M. Montiel, Jose M. and D. Tardos, Juan},
   year={2021},
   month=Dec, pages={1874–1890} }

@inproceedings{droan,
author = {Dubey, Geetesh and Madaan, Ratnesh and Scherer, Sebastian},
title = {DROAN - Disparity-Space Representation for Obstacle Avoidance: Enabling Wire Mapping \& Avoidance},
year = {2018},
publisher = {IEEE Press},
url = {https://doi.org/10.1109/IROS.2018.8593499},
doi = {10.1109/IROS.2018.8593499},
booktitle = {2018 IEEE/RSJ International Conference on Intelligent Robots and Systems (IROS)},
pages = {6311–6318},
numpages = {8},
location = {Madrid, Spain}
}

@Article{damodaran,
AUTHOR = {Damodaran, Deeptha and Mozaffari, Saeed and Alirezaee, Shahpour and Ahamed, Mohammed Jalal},
TITLE = {Experimental Analysis of the Behavior of Mirror-like Objects in LiDAR-Based Robot Navigation},
JOURNAL = {Applied Sciences},
VOLUME = {13},
YEAR = {2023},
NUMBER = {5},
ARTICLE-NUMBER = {2908},
URL = {https://www.mdpi.com/2076-3417/13/5/2908},
ISSN = {2076-3417},
DOI = {10.3390/app13052908}
}

@misc{nagata2026mirrordriftactuatedmirrorbasedattacks,
      title={MirrorDrift: Actuated Mirror-Based Attacks on LiDAR SLAM}, 
      author={Rokuto Nagata and Kenji Koide and Kazuma Ikeda and Ozora Sako and Shion Horie and Kentaro Yoshioka},
      year={2026},
      eprint={2603.11364},
      archivePrefix={arXiv},
      primaryClass={cs.RO},
      url={https://arxiv.org/abs/2603.11364}, 
}

@inproceedings{ovdb,
author = {Museth, Ken and Lait, Jeff and Johanson, John and Budsberg, Jeff and Henderson, Ron and Alden, Mihai and Cucka, Peter and Hill, David and Pearce, Andrew},
title = {OpenVDB: an open-source data structure and toolkit for high-resolution volumes},
year = {2013},
isbn = {9781450323390},
publisher = {Association for Computing Machinery},
address = {New York, NY, USA},
url = {https://doi.org/10.1145/2504435.2504454},
doi = {10.1145/2504435.2504454},
booktitle = {ACM SIGGRAPH 2013 Courses},
articleno = {19},
numpages = {1},
location = {Anaheim, California},
series = {SIGGRAPH '13}
}

@misc{oh2025languagecostproactivehazard,
      title={Language as Cost: Proactive Hazard Mapping using VLM for Robot Navigation}, 
      author={Mintaek Oh and Chan Kim and Seung-Woo Seo and Seong-Woo Kim},
      year={2025},
      eprint={2508.03138},
      archivePrefix={arXiv},
      primaryClass={cs.RO},
      url={https://arxiv.org/abs/2508.03138}, 
}

@misc{jongwiriyanurak2025vroastvisualroadassessment,
      title={V-RoAst: Visual Road Assessment. Can VLM be a Road Safety Assessor Using the iRAP Standard?}, 
      author={Natchapon Jongwiriyanurak and Zichao Zeng and June Moh Goo and Xinglei Wang and Ilya Ilyankou and Kerkritt Sriroongvikrai and Nicola Christie and Meihui Wang and Huanfa Chen and James Haworth},
      year={2025},
      eprint={2408.10872},
      archivePrefix={arXiv},
      primaryClass={cs.CV},
      url={https://arxiv.org/abs/2408.10872}, 
}

@article{Labb__2018,
   title={RTAB‐Map as an open‐source lidar and visual simultaneous localization and mapping library for large‐scale and long‐term online operation},
   volume={36},
   ISSN={1556-4967},
   url={http://dx.doi.org/10.1002/rob.21831},
   DOI={10.1002/rob.21831},
   number={2},
   journal={Journal of Field Robotics},
   publisher={Wiley},
   author={Labbé, Mathieu and Michaud, François},
   year={2018},
   month=Oct, pages={416–446} }

@misc{radio,
      title={AM-RADIO: Agglomerative Vision Foundation Model -- Reduce All Domains Into One}, 
      author={Mike Ranzinger and Greg Heinrich and Jan Kautz and Pavlo Molchanov},
      year={2024},
      eprint={2312.06709},
      archivePrefix={arXiv},
      primaryClass={cs.CV},
      url={https://arxiv.org/abs/2312.06709}, 
}

@misc{sinha2024realtimeanomalydetectionreactive,
      title={Real-Time Anomaly Detection and Reactive Planning with Large Language Models}, 
      author={Rohan Sinha and Amine Elhafsi and Christopher Agia and Matthew Foutter and Edward Schmerling and Marco Pavone},
      year={2024},
      eprint={2407.08735},
      archivePrefix={arXiv},
      primaryClass={cs.RO},
      url={https://arxiv.org/abs/2407.08735}, 
}

@misc{chakraborty2026autonomouslaboratorysafetymonitoring,
      title={Toward Autonomous Laboratory Safety Monitoring with Vision Language Models: Learning to See Hazards Through Scene Structure}, 
      author={Trishna Chakraborty and Udita Ghosh and Aldair Ernesto Gongora and Ruben Glatt and Yue Dong and Jiachen Li and Amit K. Roy-Chowdhury and Chengyu Song},
      year={2026},
      eprint={2602.00414},
      archivePrefix={arXiv},
      primaryClass={cs.CV},
      url={https://arxiv.org/abs/2602.00414}, 
}

@misc{rayfronts,
      title={RayFronts: Open-Set Semantic Ray Frontiers for Online Scene Understanding and Exploration}, 
      author={Omar Alama and Avigyan Bhattacharya and Haoyang He and Seungchan Kim and Yuheng Qiu and Wenshan Wang and Cherie Ho and Nikhil Keetha and Sebastian Scherer},
      year={2025},
      eprint={2504.06994},
      archivePrefix={arXiv},
      primaryClass={cs.RO},
      url={https://arxiv.org/abs/2504.06994}, 
}

@misc{siglip,
      title={Sigmoid Loss for Language Image Pre-Training}, 
      author={Xiaohua Zhai and Basil Mustafa and Alexander Kolesnikov and Lucas Beyer},
      year={2023},
      eprint={2303.15343},
      archivePrefix={arXiv},
      primaryClass={cs.CV},
      url={https://arxiv.org/abs/2303.15343}, 
}

@misc{siva,
      title={SALON: Self-supervised Adaptive Learning for Off-road Navigation}, 
      author={Matthew Sivaprakasam and Samuel Triest and Cherie Ho and Shubhra Aich and Jeric Lew and Isaiah Adu and Wenshan Wang and Sebastian Scherer},
      year={2024},
      eprint={2412.07826},
      archivePrefix={arXiv},
      primaryClass={cs.RO},
      url={https://arxiv.org/abs/2412.07826}, 
}

@misc{deisenroth2015distributedgaussianprocesses,
      title={Distributed Gaussian Processes}, 
      author={Marc Peter Deisenroth and Jun Wei Ng},
      year={2015},
      eprint={1502.02843},
      archivePrefix={arXiv},
      primaryClass={stat.ML},
      url={https://arxiv.org/abs/1502.02843}, 
}

@misc{long2025draedynamicretrievalaugmentedexpert,
      title={DRAE: Dynamic Retrieval-Augmented Expert Networks for Lifelong Learning and Task Adaptation in Robotics}, 
      author={Yayu Long and Kewei Chen and Long Jin and Mingsheng Shang},
      year={2025},
      eprint={2507.04661},
      archivePrefix={arXiv},
      primaryClass={cs.RO},
      url={https://arxiv.org/abs/2507.04661}, 
}

@article{OCallaghan2012GaussianPO,
  title={Gaussian process occupancy maps*},
  author={Simon Timothy O'Callaghan and Fabio Tozeto Ramos},
  journal={The International Journal of Robotics Research},
  year={2012},
  volume={31},
  pages={42 - 62},
  url={https://api.semanticscholar.org/CorpusID:34447821}
}

@article{Kocijan01122005,
author = {Juš Kocijan and Agathe Girard and Blaž Banko and Roderick Murray-Smith},
title = {Dynamic systems identification with Gaussian processes},
journal = {Mathematical and Computer Modelling of Dynamical Systems},
volume = {11},
number = {4},
pages = {411--424},
year = {2005},
publisher = {Taylor \\& Francis},
doi = {10.1080/13873950500068567},


URL = { 
    
        https://doi.org/10.1080/13873950500068567
    
    

},
eprint = { 
    
        https://doi.org/10.1080/13873950500068567
    
    

}

}

@article{Faessler_2018,
   title={Differential Flatness of Quadrotor Dynamics Subject to Rotor Drag for Accurate Tracking of High-Speed Trajectories},
   volume={3},
   ISSN={2377-3774},
   url={http://dx.doi.org/10.1109/LRA.2017.2776353},
   DOI={10.1109/lra.2017.2776353},
   number={2},
   journal={IEEE Robotics and Automation Letters},
   publisher={Institute of Electrical and Electronics Engineers (IEEE)},
   author={Faessler, Matthias and Franchi, Antonio and Scaramuzza, Davide},
   year={2018},
   month=apr, pages={620–626} }

@book{gaussianml,
    author = {Rasmussen, Carl Edward and Williams, Christopher K. I.},
    title = {Gaussian Processes for Machine Learning},
    publisher = {The MIT Press},
    year = {2005},
    month = {11},
    isbn = {9780262256834},
    doi = {10.7551/mitpress/3206.001.0001},
    url = {https://doi.org/10.7551/mitpress/3206.001.0001},
    eprint = {https://direct.mit.edu/book-pdf/2514321/book_9780262256834.pdf},
}

@article{ljung1998system,
author = {Simpkins, Charles},
year = {2012},
month = {06},
pages = {95-96},
title = {System Identification: Theory for the User, 2nd Edition (Ljung, L.; 1999) [On the Shelf]},
volume = {19},
journal = {Robotics \& Automation Magazine, IEEE},
doi = {10.1109/MRA.2012.2192817}
}

@misc{finn2017maml,
      title={Model-Agnostic Meta-Learning for Fast Adaptation of Deep Networks}, 
      author={Chelsea Finn and Pieter Abbeel and Sergey Levine},
      year={2017},
      eprint={1703.03400},
      archivePrefix={arXiv},
      primaryClass={cs.LG},
      url={https://arxiv.org/abs/1703.03400}, 
}

@misc{nagabandi2019learning,
      title={Learning to Adapt in Dynamic, Real-World Environments Through Meta-Reinforcement Learning}, 
      author={Anusha Nagabandi and Ignasi Clavera and Simin Liu and Ronald S. Fearing and Pieter Abbeel and Sergey Levine and Chelsea Finn},
      year={2019},
      eprint={1803.11347},
      archivePrefix={arXiv},
      primaryClass={cs.LG},
      url={https://arxiv.org/abs/1803.11347}, 
}

@software{NVIDIA_Isaac_Sim,
author = {{NVIDIA}},
license = {Apache-2.0},
title = {{Isaac Sim}},
url = {https://github.com/isaac-sim/IsaacSim},
version = {5.1.0}
}

@inproceedings{jacinto2024pegasus,
   title={Pegasus Simulator: An Isaac Sim Framework for Multiple Aerial Vehicles Simulation},
   url={http://dx.doi.org/10.1109/ICUAS60882.2024.10556959},
   DOI={10.1109/icuas60882.2024.10556959},
   booktitle={2024 International Conference on Unmanned Aircraft Systems (ICUAS)},
   publisher={IEEE},
   author={Jacinto, Marcelo and Pinto, João and Patrikar, Jay and Keller, John and Cunha, Rita and Scherer, Sebastian and Pascoal, António},
   year={2024},
   month=June, pages={917–922} }

@INPROCEEDINGS{meier2015px4,
  author={Meier, Lorenz and Honegger, Dominik and Pollefeys, Marc},
  booktitle={2015 IEEE International Conference on Robotics and Automation (ICRA)}, 
  title={PX4: A node-based multithreaded open source robotics framework for deeply embedded platforms}, 
  year={2015},
  volume={},
  number={},
  pages={6235-6240},
  doi={10.1109/ICRA.2015.7140074}}

\end{document}